\documentclass[conference]{IEEEtran}
\IEEEoverridecommandlockouts
\usepackage{cite}
\usepackage{amsmath,amssymb,amsfonts}
\usepackage{algorithmic}
\usepackage{graphicx}
\usepackage{textcomp}
\usepackage{xcolor}
\def\BibTeX{{\rm B\kern-.05em{\sc i\kern-.025em b}\kern-.08em
    T\kern-.1667em\lower.7ex\hbox{E}\kern-.125emX}}

\makeatletter
    \let\MYcaption\@makecaption
\makeatother

\usepackage[font=footnotesize, labelformat=simple]{subcaption}

\newcommand{\RNum}[1]{\uppercase\expandafter{\romannumeral #1\relax}} 

\begin{document}

\title{Towards end-to-end pulsed eddy current classification and regression with CNN
}

\author{\IEEEauthorblockN{Xin Fu}
\IEEEauthorblockA{\textit{School of Electronic Information} \\
\textit{Wuhan University}\\
Wuhan, China \\
fuxin@whu.edu.cn}
\and
\IEEEauthorblockN{Zheng Liu, Chengkai Zhang, Xiang Peng, Lihua Jian}
\IEEEauthorblockA{\textit{School of Engineering} \\
\textit{University of British Columbia}\\
Kelowna, Canada \\
zheng.liu@ubc.ca}
}

\maketitle

\begin{abstract}
    Pulsed eddy current (PEC) is an effective electromagnetic non-destructive inspection (NDI) technique for metal materials, which has already been widely adopted in detecting cracking and corrosion in some multi-layer structures.
    Automatically inspecting the defects in these structures would be conducive to further analysis and treatment of them.
    In this paper, we propose an effective end-to-end model using convolutional neural networks (CNN) to learn effective features from PEC data. Specifically, we construct a multi-task generic model, based on 1D CNN, to predict both the class and depth of flaws simultaneously. Extensive experiments demonstrate our model is capable of handling both classification and regression tasks on PEC data. Our proposed model obtains higher accuracy and lower error compared to other standard methods.
\end{abstract}


\section{Introduction}
\begin{figure}[t]
  \centering
  \includegraphics[width=0.6\linewidth]{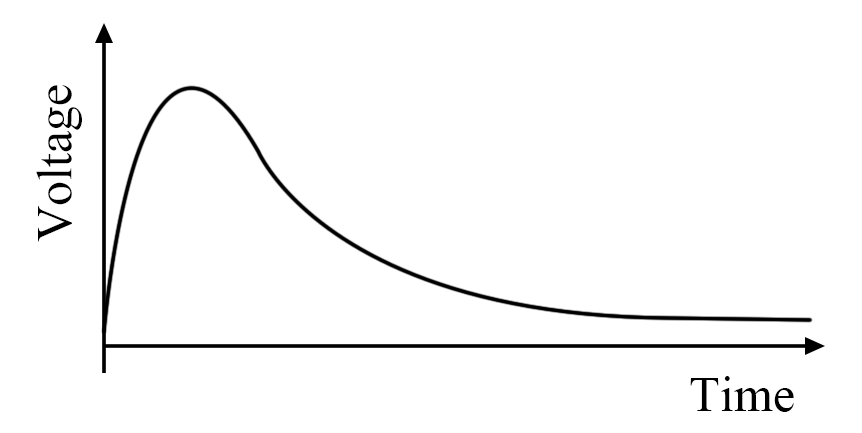}
  \caption{A typical pulsed eddy current time-domain response signal.}
  \label{fig:typical-pec}
\end{figure}

Pulsed eddy current (PEC) is a new emerging non-destructive inspection (NDI) technique in recent decades. It exploits a step function voltage to excite the probe. Typically, a PEC signal waveform is a time series with the pattern shown in Figure \ref{fig:typical-pec}, where the current signal decays and approaches a steady state eventually. It has two major features: peak value and zero crossover point \cite{BieberTimeGatingPulsedEddy1997}, which are mainly used in flaw characterization. Waveforms are acquired continuously while the probe is scanned over an area of the sample using a portable two-axis scanner.
PEC has a wide range of frequencies, both surface and subsurface flaws of metal material can be detected. Therefore, PEC technique has been broadly utilized in the practices of non-destructive testing and evaluations \cite{SophianPulsedEddyCurrent2017} \cite{StottPulsedEddyCurrent2015} \cite{HePulsededdycurrent2010} \cite{PanPECFrequencyBand2013} \cite{GaoAutomaticDefectIdentification2014}. 
Hence, automatically classifying and analyzing PEC signals have become a challenging problem for the research community.

So far, quite a few researchers have developed ways to characterize and quantify the corrosion based on PEC technique. Reference \cite{LiuInvestigationsclassifyingpulsed2003a} extracts a few features and uses MLP to classify them. 
In literature \cite{HeSupportvectormachine2013}, principal component analysis (PCA) and independent component analysis have been investigated for defect classification in multilayer aluminum structures. It uses support vector machines (SVMs) \cite{HearstSupportVectorMachines1998} as the classifier.
In research work \cite{BuckSimultaneousMultiparameterMeasurement2016}, a modified PCA (MPCA) is proposed to process the input data. Then an artificial neural network is created to give the prediction.
Additionally, some researchers have investigated different ways to classify NDI signals by neural network \cite{BuckSimultaneousMultiparameterMeasurement2016} \cite{LiuInvestigationsclassifyingpulsed2003a} \cite{PredaNeuralnetworkinverse1999}. 
The major steps of them include: signal preprocessing, feature extraction and classification. 

\begin{figure*}[t]
  \begin{center}
      \includegraphics[width=0.7\linewidth]{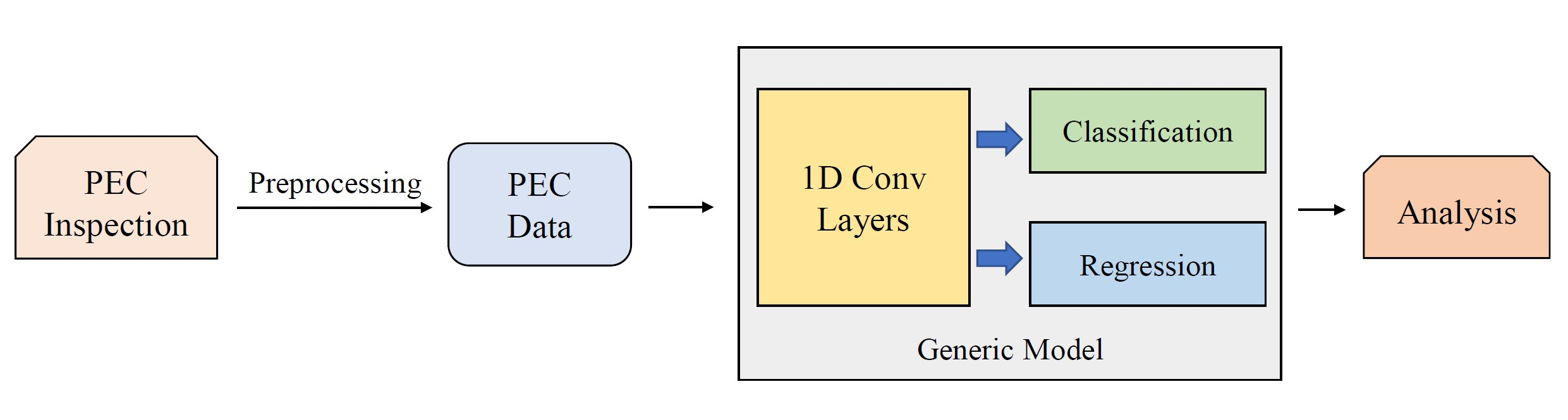}
  \end{center}
  \caption{The framework of our proposed method.}
  \label{fig:method-flow}
\end{figure*}

Nonetheless, these schemes heavily rely on manually chosen features rather than learned features. The handcrafted features are too crucial to the final performance and obviously have severe limitations. Moreover, most of the schemes are single-task oriented, which means they are mainly intended for one task.
In some real applications, we need to classify the collected PEC signals, and we also want to get approximate estimation of the metal loss. 
Thus, a robust and multi-task model is needed, which doesn't need prior chosen features and can tackle different tasks at the same time. 

In recent years, deep learning has developed intensively and been adopted in a wide range of research fields \cite{Goodfellow-et-al-2016}. Among the many successful deep neural networks, convolutional neural network (CNN) has excellent performance in different tasks such as image classification and sequence modeling \cite{LeCunBackpropagationAppliedHandwritten1989} \cite{BaiEmpiricalEvaluationGeneric2018} \cite{KimConvolutionalNeuralNetworks2014} \cite{ReneTemporalConvolutional2017}.
Basically, sliding kernels and filters are utilized by CNNs to create their own feature maps. These filters have trainable weights, each of which can be trained through back-propagation. By stacking the layers of convolutional filters together, we can get hierarchical feature maps. They allow the network to operate on unprocessed features, reducing the need for handcrafted features or feature engineering. 
Thus, theoretically, CNN layers can avoid the shortcomings from the features chosen manually. In addition, they can enhance the result by exploiting potential features learned from the data.

A majority of the previous research work uses shallow neural network like multilayer perceptron (MLP) on PEC data. Since they haven't explored the potentials of CNNs, we want to take a leap by exploiting the state-of-the-art CNN technique in problems with respect to PEC signals.
Plus, we would also like to introduce multi-task learning technique \cite{RuderOverviewMultiTaskLearning2017} into our method, which aims to learn several tasks simultaneously to boost the performance of the main task or all tasks. 

In this paper, we propose an effective end-to-end model based on the latest CNN architecture, which is designed to tackle the classification and regression task of the PEC inspection data simultaneously. Compared to conventional methods that use handcrafted features and MLP, our proposed model directly takes the signal as the input and yields the straightforward results using CNN. Experiments performed on PEC data from real-life specimens demonstrate that our model is capable of accurately tackling the classification and regression task of the PEC inspection data simultaneously.

\section{Methodology}

The entire pipeline based on our proposed model is depicted in Figure \ref{fig:method-flow}. There are mainly three stages in the flow: (i) preprocessing the data; (ii) feeding the data to the model; and (iii) analyzing the outputs.

Preprocessing is meant to remove the redundant parts and augment the original data. Analyzing the outputs should be designed according to the real application scenario. 
The following part gives a detailed explanation of our proposed model.

\begin{figure}[h]
  \centering
  \includegraphics[width=0.7\linewidth]{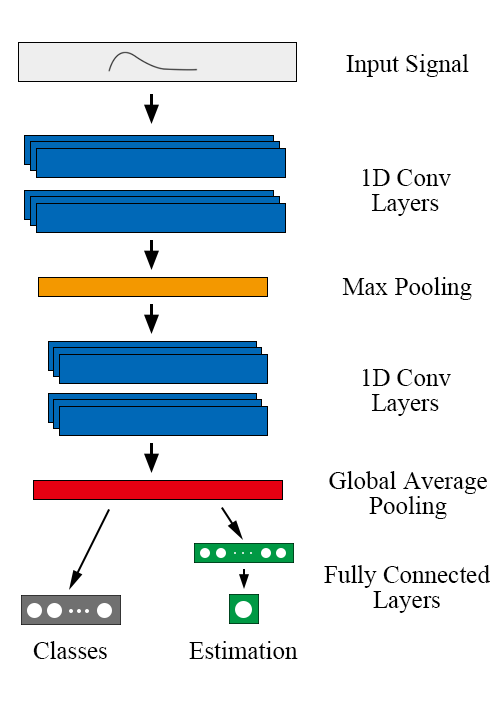}
  \caption{Our proposed generic model based on CNN.}
  \label{fig:generic-model}
\end{figure}

\subsection{End-to-end Multi-task CNN}
\label{sec:model}
\subsubsection{Generic model}

Same as conventional signal convolution operation, 1-dimensional convolution (1D Conv) layer is introduced in modern neural networks to process sequential signal \cite{LecunConvolutional1995}. The 1D Conv layer can be regarded as a fuzzy filter that can enhance the features of original signals and reduce noises. PEC signals are temporal and have fixed width at each inspection point. Therefore, 1D Conv layer is adopted in our proposed model. 

Pooling layers are commonly used for non-linear down-sampling in CNN models. In our proposed model, we adopt max pooling layer, which uses the maximum value from each of a cluster of activations at the prior convolutional layer.
To increase the efficiency of our model, global average pooling (GAP) \cite{LinNetworkNetwork2013} layer will be exploited to reduce the dimension of the outputs from the last convolutional layer.

The first section serves as a powerful feature extractor. It can capture the implicit information and effective features from PEC signals. As is shown in Figure \ref{fig:generic-model}, it's made of several parts:
(i) 1D Conv Layers,
(ii) Max Pooling Layer, 
(iii) 1D Conv Layers, 
(iv) Global Average Pooling Layer.

Since the first part intakes the PEC signals and outputs effective latent feature vectors, we construct the second part according to specific tasks. For classification, we use fully connected (FC) layer with softmax activation, which takes the output of the GAP layer and produce a categorical distribution. Meanwhile, for the regression problem, which is useful for predicting the depths of corrosion, two FC layers are used. The first layer is directly connected to the prior GAP layer. The second layer has only one output, and the activation function is removed so that it gives continuous output values.

Thus, unlike the previously proposed single-task methods, our generic model can not only handle the classification, but also the regression problem. It would be extremely helpful in different scenarios of PEC data analysis.
The entire model is illustrated in Figure \ref{fig:generic-model}. Since this is a generic model, we are able to design and tweak the parameters based on the PEC data obtained from different scenarios. 

\subsubsection{Loss Function}

Choosing suitable loss functions are vital to the optimization and convergence of our multi-task model. 

So far, there are a variety of loss functions for classification such as $\mathcal{L}_1$, $\mathcal{L}_2$, margin loss and log loss \cite{JanochaLossFunctionsDeep2017}. 
For the classification part, following conventional approaches, we select categorical cross entropy loss. The formula of cross entropy loss is shown in Equation \ref{eq:loss_c},

\begin{equation} 
  \label{eq:loss_c}
  \mathcal{L}_c = -\frac{1}{N}\sum_{i=1}^{N}\sum_{c=1}^{M}{y_{i,c} log(p_{i,c})}
\end{equation}

In the above equation, $i$ and $c$ index observations and classes, $N$ and $M$ indicate the total number of samples and categories, $y_{i,c}$ denotes the binary indicator for observation, and $p_{i,c}$ is the probability prediction satisfying $p_{i,c}\in[0,1], \sum_{c} p_{i,c} =1$.

For the regression part, we choose mean absolute error (MAE) as the loss function, as shown in Equation \ref{eq:loss_r}. For sample $i$, $y_i$ represents the label value and $\hat{y_i}$ is the output from the model.

\begin{equation} 
  \label{eq:loss_r}
  \mathcal{L}_r = \frac{1}{N}\sum_{i=1}^{N}{ |y_i - \hat{y_i}| }
\end{equation}

Thus, the combined loss, as shown in Equation \ref{eq:loss_all}, will be used during the training of our proposed multi-task learning model. 

\begin{equation} 
  \label{eq:loss_all}
  \begin{split}
    \mathcal{L}_{all} & = \alpha \mathcal{L}_c + \beta \mathcal{L}_r \\
    & =  -\frac{\alpha}{N}\sum_{i=1}^{N}\sum_{c=1}^{M}{y_{i,c} log(p_{i,c})} + \frac{\beta}{N}\sum_{i=1}^{N}{ |y_i - \hat{y_i}| }
  \end{split}
\end{equation}

The parameters $\alpha, \beta$ should be set according to different scenarios to adjust the balance between classification and regression losses.

\section{Experiments}
\label{sec:experiments}

\subsection{Experimental Setup and Specimen}

In this paper, we mainly carry out experiments on two specimens. They are denoted as Specimen A and Specimen B. 

\begin{figure}[t]
    \centering
    \begin{subfigure}[b]{0.45\textwidth}
      \centering
      \includegraphics[width=0.7\linewidth]{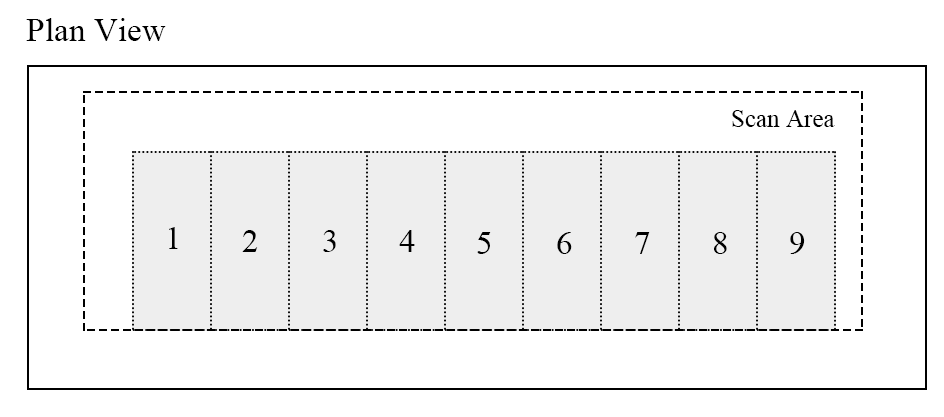}
      \caption{}
      \label{fig:SpecimenA-a} 
    \end{subfigure}
    \begin{subfigure}[b]{0.45\textwidth}
      \centering
      \includegraphics[width=0.7\linewidth]{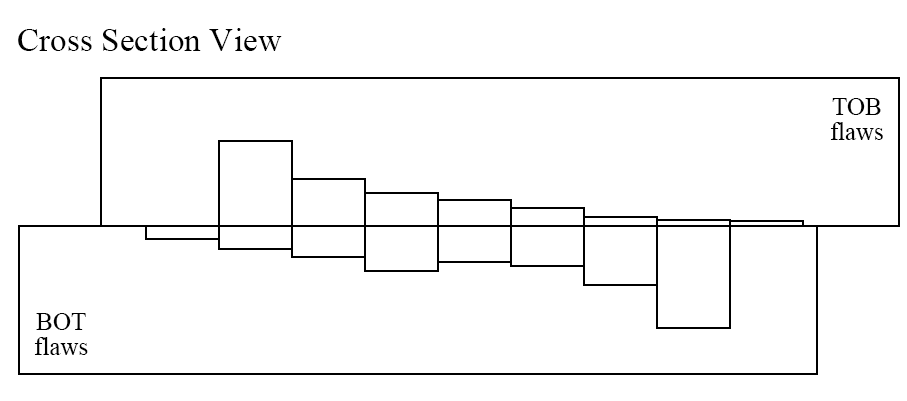}
      \caption{}
      \label{fig:SpecimenA-b} 
    \end{subfigure}
    \begin{subfigure}[b]{0.45\textwidth}
      \centering
      \includegraphics[width=0.7\linewidth]{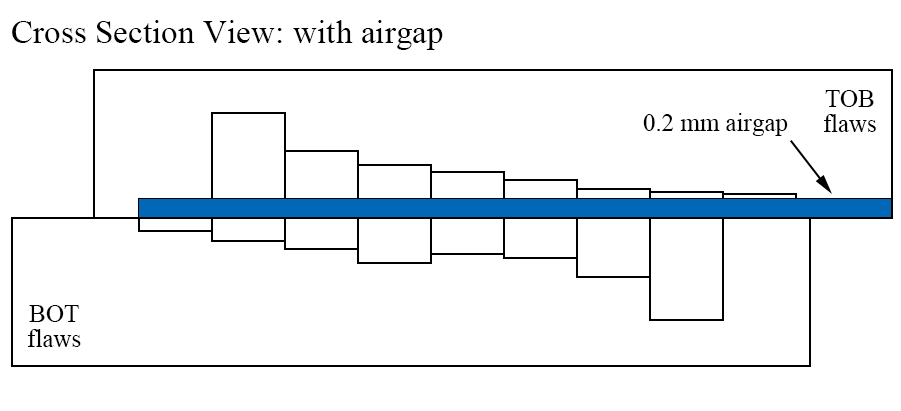}
      \caption{}
      \label{fig:SpecimenA-c} 
    \end{subfigure}
    \caption{Schematic of Specimen A: (a) plan view of the specimen; (b) cross-section view of specimen without airgap; and (c) cross-section view of specimen with airgap.}
    \label{fig:SpecimenA}
\end{figure}

\subsubsection{Specimen A}

This piece of specimen, shown in Figure \ref{fig:SpecimenA}, originally comes from literature \cite{LiuInvestigationsclassifyingpulsed2003a}. It was assembled by two 1 mm-thick sheets of Al2024 T3, which have regions milled out to simulate material loss due to corrosion.  
The flaws are composed of two parts: top of bottom (TOB) layer and bottom of top (BOT) layer. Each layer has 8 places that have been milled out a certain depth of material, as is shown in Figure \ref{fig:SpecimenA-b}.

There are nine levels of varying material loss and location in this specimen. The flaws for the two parts are (in mm): (i) BOT: 0.4826, 0.2519, 0.1884, 0.1545, 0.1143, 0.0572, 0.0402, 0.021; (ii) TOB: 0.0847, 0.1431, 0.1693, 0.2032, 0.1736, 0.1820, 0.2413, 0.4826. 

A section of the test specimen was inspected at a resolution of 1 mm in x and y directions, resulting in a 65 pixel by 250 pixel area as shown in Fig. \ref{fig:SpecimenA-a}. Besides, a second inspection is carried out after an airgap is introduced, as shown in Fig. \ref{fig:SpecimenA-c}. 

For Specimen A, PEC data collected from the sections with metal loss are equally and randomly selected for training and testing. For each area which has $1250(50\times25)$ data points, $625$ of them are used as training and testing set, respectively. From the loss-free area, we selected an area with shape $5\times250$ from the upper part of the specimen whose y-axis ranges from $55$ to $60$. Also, it has $1250$ samples and will be equally partitioned as former sections. Therefore, the training and testing sets both have $10$ classes with total $6250(10\times625)$ samples.
After the airgap is applied to Specimen A, we also use the same method as described above to yield the training and testing sets for our experiment.

For clarification, the dataset collected from Specimen A without airgap is denoted as A\textsubscript{1}, and the set from Specimen A with airgap is denoted as A\textsubscript{2}. Besides, by combining A\textsubscript{1} and A\textsubscript{2} together, we could get a larger set called A\textsubscript{1}+A\textsubscript{2}, which includes both of the data with and without airgap.


\begin{figure}[t]
  \centering
  \begin{subfigure}[b]{0.45\textwidth}
    \centering
    \includegraphics[width=0.8\linewidth]{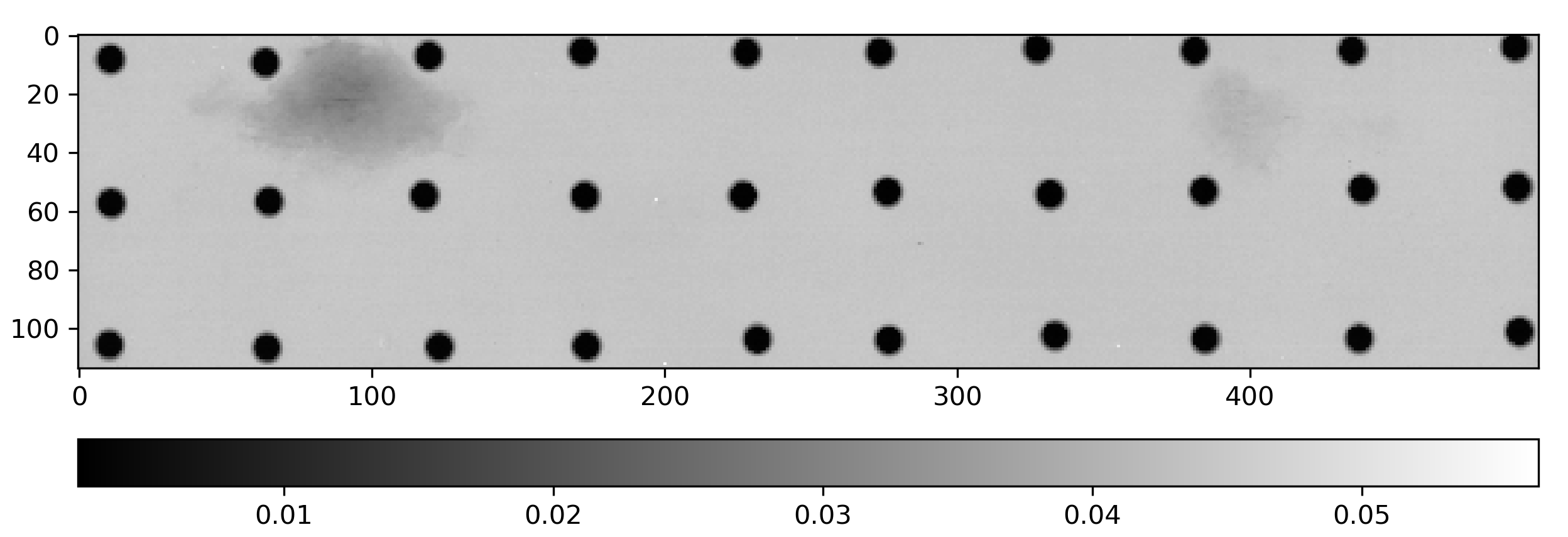}
    \caption{}
    \label{fig:SpecimenB-2} 
  \end{subfigure}
  \begin{subfigure}[b]{0.45\textwidth}
    \centering
    \includegraphics[width=0.8\linewidth]{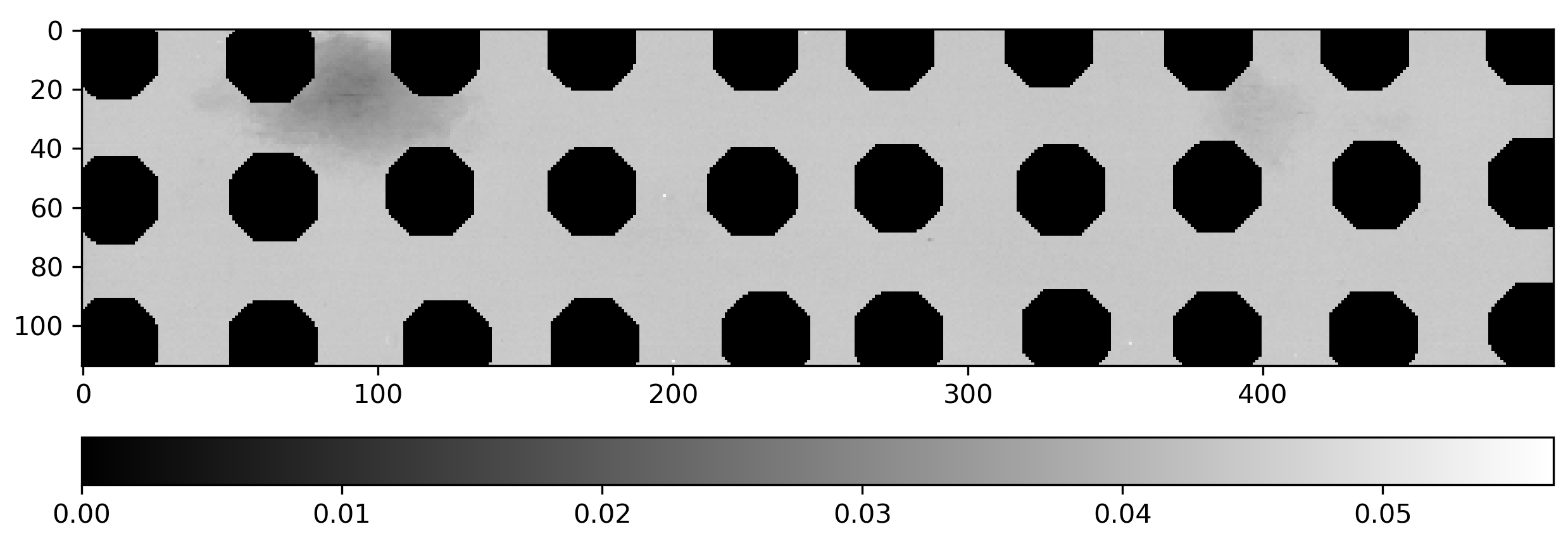}
    \caption{}
    \label{fig:SpecimenB-3} 
  \end{subfigure}
  \caption{X-ray thickness map of Specimen B: (a) original scanned specimen; (b) after removing the riveted parts.}
  \label{fig:SpecimenB}
\end{figure}

\subsubsection{Specimen B}

The second specimen in our experiment is used in \cite{LiuDataFusionSchemeQuantitative2008}. Different from Specimen A, which is made with artificial metal loss, Specimen B is from an actual aircraft lap joint. As Figure \ref{fig:SpecimenB} shows, it's from a service-retired Boeing 727. By using this piece of material, we are intended to investigate the effectiveness of our proposed method on PEC data from real industrial scenarios.

Specifically, the PEC data of Specimen B is collected from two sections: \textit{D} and \textit{C} of the lap joint. Section \textit{D} and \textit{C} are used for training and testing respectively. Section \textit{D} has a shape of $114 \times 366$, and \textit{C} is $114 \times 499$. For each point at the section, the PEC signal is a $1\times100$ vector, which means the PEC inspection of each point has $100$ timesteps.

A sample X-ray thickness map of section \textit{C} is given in Figure \ref{fig:SpecimenB}, which serves as the ground truth label used in our training and evaluation of the model.
As we can see in Figure \ref{fig:SpecimenB-2}, the ``black holes" are the riveted sections of the lap joint, which should be removed. We set the threshold to 0.05 and mark the areas whose value is smaller than the threshold. Then we dilate the areas with size of 12 and remove them. The remaining part of the specimen will be used for training and testing, as shown in Fig. \ref{fig:SpecimenB-3}.
After removing the rivets, there will be 27,397 training samples and 37,603 samples for testing.

\subsection{Experimental Results}

The models used in our following experiments are constructed based on the generic model described in Section \ref{sec:model}. 

Specifically, for Specimen A, we set our model as follows: the first two 1D convolutional layers both have $128 \times 3\times1$ kernels and the second two have $64 \times 3 \times1$ kernels. The pooling size in the max pooling layer is $3\times1$. The output of the GAP layer would be $64\times1$.
In the classification task, a fully connected layer with 10 units will take the outputs from the GAP layer and makes the categorical prediction. In the regression task, two FC layers will be adopted, where the first follows the GAP layer with 32 units, the second follows the previous FC layer with 2 units which gives the estimation of BoT and ToB flaws.
The loss follows the Equation \ref{eq:loss_all} with $\alpha=\beta=1$. We use Adam \cite{KingmaAdamMethodStochastic2014} as the optimizer of our model. The learning rate is set to 0.001 and $\beta_1, \beta_2$ to be 0.9 and 0.999 respectively.

\begin{figure}[t]
  \begin{minipage}[b]{0.48\linewidth}
    \centering
    \centerline{\includegraphics[height=0.7\linewidth]{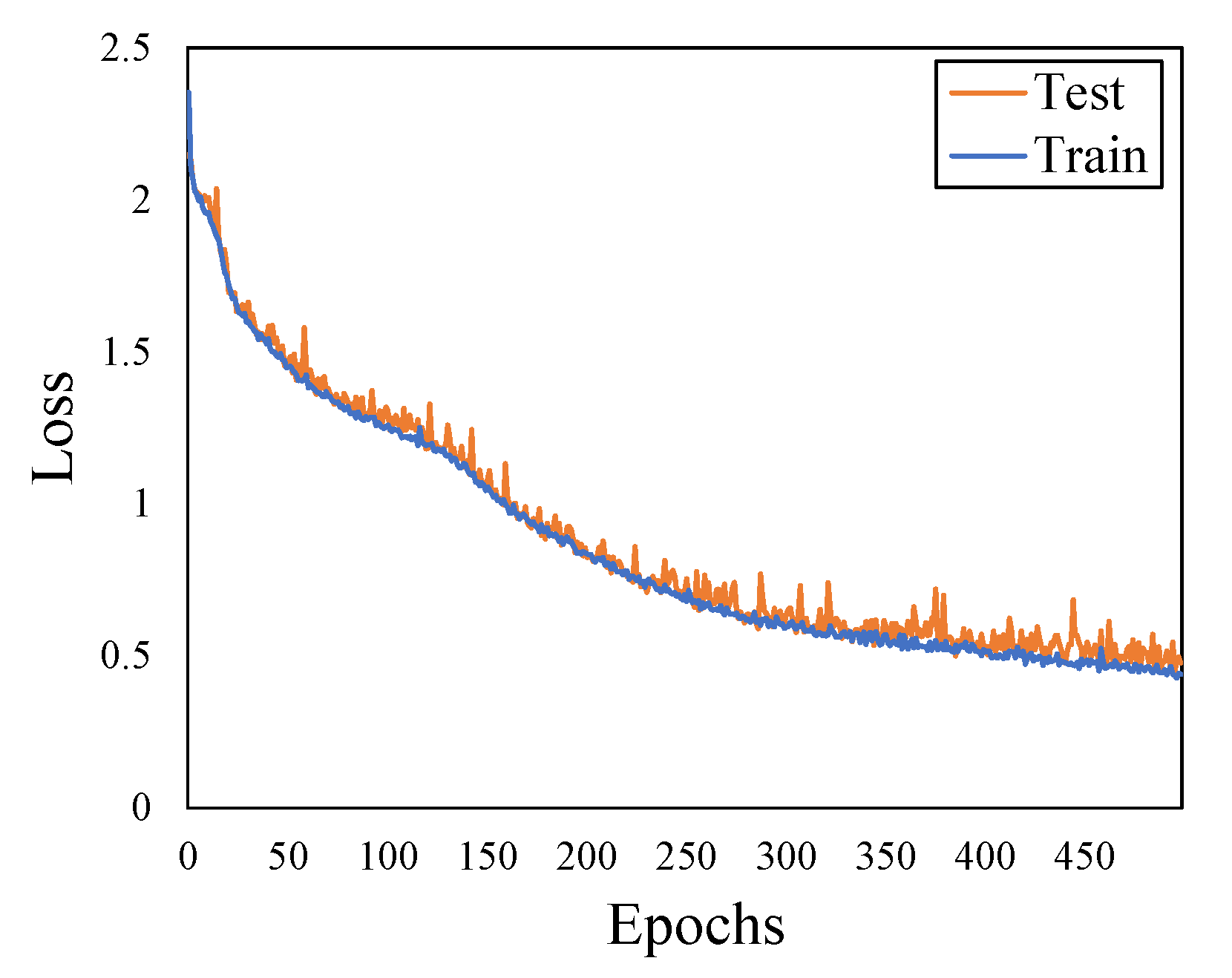}}
    \centerline{(a)}\medskip
  \end{minipage}
  \hfill
  \begin{minipage}[b]{0.48\linewidth}
    \centering
    \centerline{\includegraphics[height=0.7\linewidth]{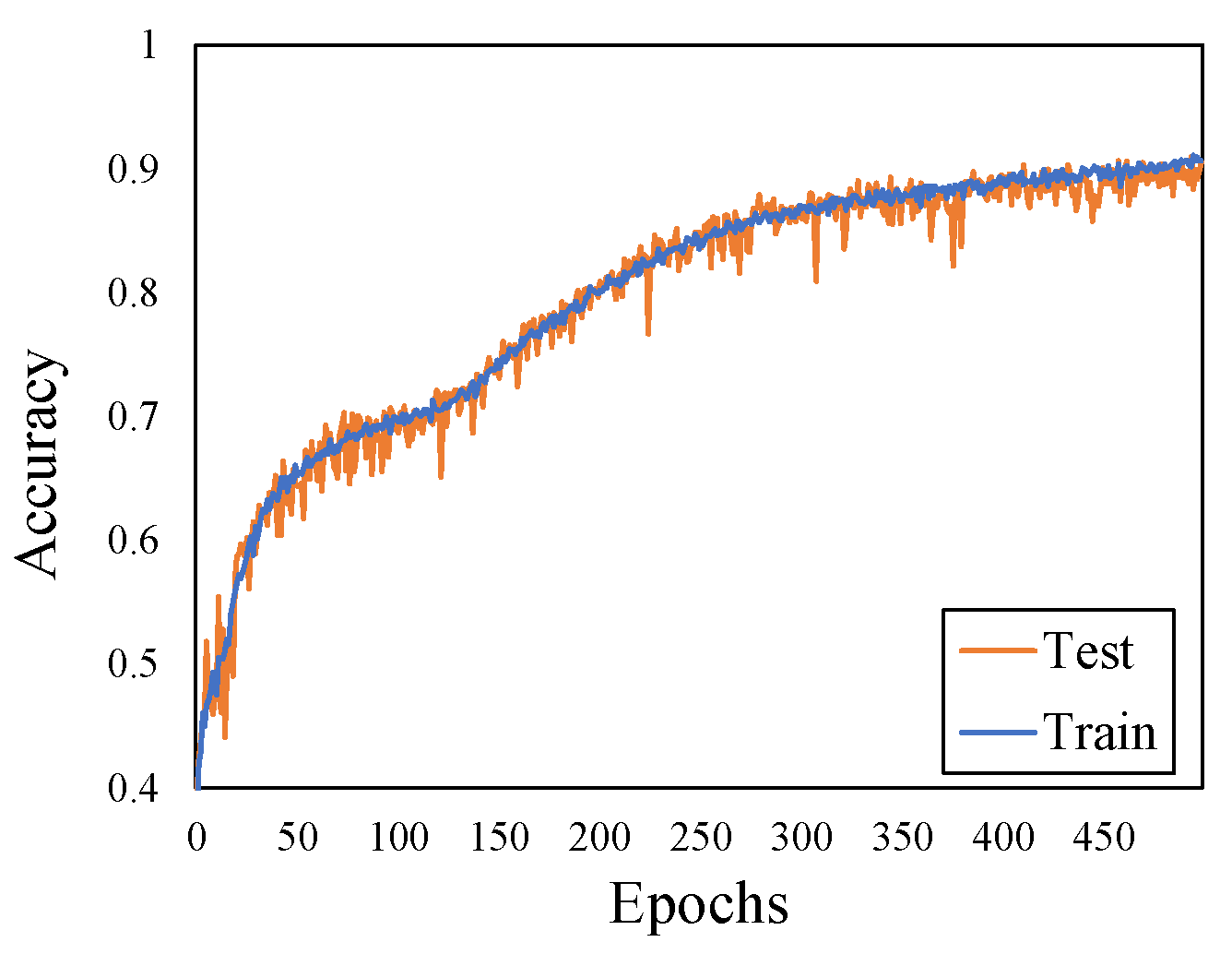}}
    \centerline{(b)}\medskip
  \end{minipage}
  \caption{Loss and accuracy curve for training and testing on Specimen A without airgap: (a) loss, (b) accuracy.}
  \label{fig:loss-acc}
  \end{figure}

The training data have 10 classes and these categorical labels are converted into one hot encoding for the sake of calculating the cross entropy loss in Equation \ref{eq:loss_c}.
The ground truth values of metal loss are relatively small, so we scale the labels of metal loss by a factor of 10.
We train our model for 500 epochs. The loss and accuracy curve for the classification is displayed in Figure \ref{fig:loss-acc}. It shows that our proposed model can converge on the PEC data from Specimen A. 

As we can see from the final results, the classification accuracy is over $90\%$. Plus, the mean squared error (MSE) of the estimation is really small. The results are shown in Table \ref{tab:results-a}.
Moreover, we compare our method with some other classification methods: \textit{Gaussian Process} \cite{RasmussenGaussianProcesses2004}, \textit{Decision Tree} \cite{SafavianSurveyDecisionTree1991} and \textit{Support Vector Machine}. Our model outperforms other classifiers on dataset A\textsubscript{1} with 91\% accuracy.

\begin{table}[t]
  \renewcommand{\arraystretch}{1.3}
  \caption{Results on Specimen A}
  \label{tab:results-a}
  \centering
  \begin{tabular}{c|c|c}
      \hline
      Dataset & \multicolumn{1}{p{2cm}|}{\centering Classification \\ Accuracy (\%)} & \multicolumn{1}{p{2cm}}{\centering Regression \\ Error (MSE)}\\
      \hline
      \hline
      A\textsubscript{1} (no airgap) & 91.0 & 0.204\\
      \hline
      A\textsubscript{2} (with airgap) & 91.2 & 0.902\\
      \hline
      A\textsubscript{1}+A\textsubscript{2} & 89.85 & 1.120\\
      \hline
  \end{tabular}
  \end{table}

\begin{table}[t]
  \caption{Comparison with other methods on A\textsubscript{1}}
  \label{tab:compare-a}
  \centering
  \begin{tabular}{c|c}
      \hline
      Method & Accuracy (\%)\\
      \hline \hline
      Gaussian Process & 53.6 \\ \hline
      Decision Tree & 87.2\\ \hline
      Support Vector Machine & 61.9\\ \hline
      \textbf{Ours} & \textbf{91.0}\\ \hline
  \end{tabular}
\end{table}

\begin{table}[t]
  \caption{Results on Specimen B}
  \label{tab:results-b}
  \centering
  \begin{tabular}{c|c}
      \hline
      Method & Mean Squared Error (MSE)\\
      \hline
      \hline
      Linear Regression & 1.133 \\
      \hline
      Bayesian Ridge Regression & 1.098\\
      \hline
      Support Vector Regression & 0.841\\
      \hline
      \textbf{Ours} & \textbf{0.498}\\
      \hline
  \end{tabular}
\end{table}

The task for Specimen B is regression. Thus, we keep the parameters as same as the model built for Specimen A. The only modification we do is to change the final FC layer to 1 unit. The X-ray thickness value, which ranges from 0.01 to 0.05, is really small. To get better regression result, we enlarge the label value by 1000. Then we start training the model for 100 epochs on data from Specimen B. 

Testing result is shown in Table \ref{tab:results-b}, that the MSE of our CNN-based model is 0.498.
We also conduct the regression task using some traditional regression methods, such as \textit{Linear Regression} \cite{NeterAppliedLinearStatistical1996}, \textit{Bayesian Ridge Regression} \cite{HoerlRidgeRegression1970} and \textit{Support Vector Regression} \cite{BasakSupportVectorRegression2007}, and the results are in Table \ref{tab:results-b}. It's obvious that our model is more robust and the estimation error is smaller than other regression methods.

\subsection{Analysis and Discussion}

From above experiments, our proposed model shows effectiveness in both tasks for classification and regression. It achieves better results on two specimens than some traditional machine learning techniques.
Hence, the proposed 1D CNN-based model can effectively capture the features from PEC data. These features could be used to predict the classes and give the estimation of the metal loss, which may be very useful in some situations for experts to determine the type of the flaws from multilayer structure.

We've done extra experiments and discovered that our model is sensitive to airgap. The result shows that the model trained on the data from specimen without airgap could not recognize the signals from specimen with airgap, and vice versa.
A possible explanation is: our model is data-driven so the features are learned from the training data. Since the model is only trained on the data distribution without airgap, as a result, testing on a distribution with airgap would be hard.
To resolve this problem, we could get as much data as possible with different levels of airgap. By tweaking and training the model on the entire data, it will be robust enough to classify the type of the metal loss from PEC signals.

\section{Conclusion}

In this paper, we propose an effective end-to-end and multi-task model based on 1D CNN to tackle the classification and regression task for PEC signals. Instead of using handcrafted features, our model can automatically learn effective features from the data, which can be used for two tasks simultaneously. Experiments on two specimens demonstrate that our model is very robust and has better performance than other methods on PEC data. Our proposed method has potentials in industrial scenarios to assist experts on inspecting the defects from collected PEC data.



\bibliographystyle{IEEEtran}
\bibliography{ref}

\vspace{12pt}

\end{document}